# Information/Relevance Influence Diagrams

Ali Jenzarli
Department of Management
The University of Tampa
401 West Kennedy Bvd. Box 15F
Tampa, FL 33606-1490, USA
(813) 253-3333, ajenzarl@cfrvm.cfr.usf.edu

## Abstract

In this paper we extend the influence diagram (ID) representation for decisions under uncertainty. In the standard ID, arrows into a decision node are only informational; they do not represent constraints on what the decision maker can do. We can represent such constraints only indirectly, using arrows to the children of the decision and sometimes adding more variables to the influence diagram, thus making the ID more complicated. Users of influence diagrams often want to represent constraints by arrows into decision nodes. We represent constraints on decisions by allowing relevance arrows into decisions nodes. We call the resulting representation *information/relevance influence diagrams* (IRIDs). Information/ relevance influence diagrams allow for direct representation and specification of constrained decisions. We use a combination of stochastic dynamic programming and Gibbs sampling to solve IRIDs. This method is especially useful when exact methods for solving IDs fail.

## 1 INTRODUCTION

In this paper we introduce a variation on influence diagrams that allows the use of arrows into decision nodes to represent constraints as well as information. We call this new representation an *Information / Relevance Influence Diagram (IRID)*. In an IRID, we model a decision variable that is constrained by other variables by drawing relevance arrows from the other variables to the decision variable. But we also still allow purely informational arrows into decision nodes.

Information/relevance influence diagrams allow for direct representation and specification of constrained decisions. This representation is a partial solution to the broader problem of representing and solving asymmetric decision problems. Asymmetry in decision problems occurs when some values of decision and/or chance variables are not allowed given certain values of their predecessors. Recent works by Smith et al. (1993), Shenoy (1993) and all the references therein offer a more complete treatment of this problem. However, none of them provide a direct and explicit representation of asymmetries concerning decision variables.

The advantage of our representation lies in the solution method we use to evaluate IRIDs. This method is a combination of stochastic dynamic programming and Gibbs sampling, an iterative Markov chain Monte Carlo algorithm (Jenzarli, 1995). This method allows us to model asymmetries concerning chance variables by using zero-one conditionals. Our solution method is especially useful when exact methods for solving influence diagrams fail.

In this paper we assume that the reader is already familiar with directed acyclic graphs (DAGs), and belief networks (Pearl, 1988). Briefly, we define a BN as a DAG in which nodes represent variables, together with a specification for each variable of a conditional probability distribution for that variable given its parents. (If there are no parents, this is a marginal probability distribution for the variable.) It is assumed that the joint probability distribution of the variables is the product of these conditional probability distributions.

We also assume that the reader is already familiar with Markov chain Monte Carlo (Hastings, 1970) and Gibbs sampling (Geman and Geman, 1984; and Gelfand and Smith, 1990).

We organize this paper as follows. In Section 2 we review influence diagrams (IDs) and their properties. In Section 3 we describe solutions algorithms for IDs. In Section 4 we describe information/relevance influence diagrams. In Section 5 we adapt the Gibbs sampling algorithm of Jenzarli (1995) to IRIDs and illustrate with an example.

## 2 INFLUENCE DIAGRAMS

In this section we review influence diagrams (IDs) and their properties. We describe the assumptions on which the ID decision model is based. And, we conclude with a note on randomization. We use a numerical example to illustrate ideas and concepts where appropriate.

Let us begin by recalling the standard definition of an influence diagram.

**Definition 1.** An *influence diagram (ID)* is a DAG with variables as nodes, together with a specification for only *some* of the variables of conditionals given their parents. We call those variables with conditionals *chance* variables, and those without *decision* variables. For each chance variable, we specify a set of *possible values*. For each decision variable, we specify a set of admissible values that we call *decision alternatives*. An ID has a special chance node that is a sink and that is a deterministic function of its parents. We call this node the *value* node. Arrows into chance nodes are called *relevance arrows* while arrows into decision nodes are called *informational arrows*.

Consider the oil wildcatter's problem adapted from Raiffa (1968). The wildcatter has to decide whether to drill or not drill. This decision is represented by the decision variable D whose values are d (drill) and nd (not drill). The values d and nd are decision alternatives. The wildcatter is uncertain about the amount of oil in the well. The amount of oil in the well is represented by the chance variable O, whose possible values are w (wet) and y (dry). The cost of drilling is $950,000, and the estimated revenue from drilling a wet well is $2,000,000. Table 1 shows the wildcatter's net monetary payoffs and his subjective probabilities for O.

**Table 1.** Payoff table for the wildcatter's problem

| | | D | | |
|---|---|---|---|---|
| State | | d (drill) | nd (not drill) | Probability of State |
| O | y (dry) | –$950,000 | 0 | 0.4 |
| | w (wet) | $1,050,000 | $0 | 0.6 |

More information about the amount of oil in the well can be obtained by conducting one of two seismic tests; test1 is standard and relatively affordable, while test2 is advanced and more costly. The seismic testing decision is represented by the decision variable T, whose values are t1 (test1), t2 (test2) and nt (no test). The seismic test result is represented by the chance variable R, whose values are o (open structure), supporting a dry well, c (closed structure), supporting a wet well, and nr (no result), indicating that no seismic test has been performed. Finally, the payoff function is represented by the value node V, which is a function of O, T and D.

Figure 1 shows an ID for the oil wildcatter's problem, where chance variables are shown as circles, decision variables as rectangles, and the value node as a diamond.

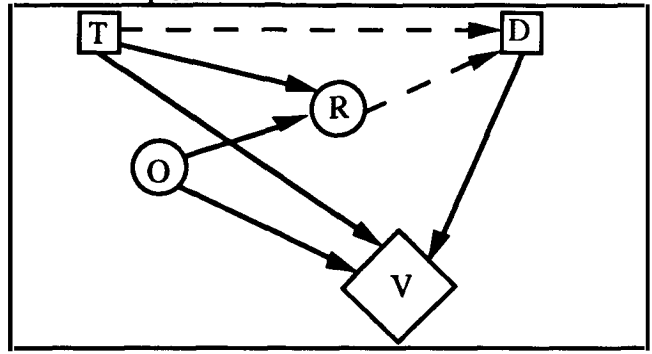

**Figure 1.** Influence diagram for the oil wildcatter's problem

Chance variables in an ID are interpreted just as they are in a BN. The conditional probability

$$P(X = x | X_1 = x_1, X_2 = x_2, ..., X_k = x_k) \quad (1)$$

that we specify for a chance variable X indicates our belief that X will take the value x when its parents, the $X_i$, take the values $x_i$. For the sake of generality, we will not rule out the possibility that the values of the conditional probability are unaffected as we change some of the $x_i$. By drawing arrows to X from all the $X_i$ we signal that we think X depends on them all, and that we expect to specify values for (1) that do depend on them all. But if assessment of the probabilities follows the construction of the DAG, this expectation may fail to be fulfilled. In Figure 1 the chance variable R depends on the values of the decision variable T and the chance variable O. The conditional probability distribution of R given T and O is shown in Table 2. Notice that the conditional probability of R given T = nt and O remains unchanged whether O = w or O = y.

**Table 2.** Conditional probability distribution of R given T and O

| T | O | R | | |
|---|---|---|---|---|
| | | o | c | nr |
| t1 | w | 0.2 | 0.8 | 0 |
| t1 | y | 0.9 | 0.1 | 0 |
| t2 | w | 0.05 | 0.95 | 0 |
| t2 | y | 0.95 | 0.05 | 0 |
| nt | w | 0 | 0 | 1 |
| nt | y | 0 | 0 | 1 |

The decision maker can be asked to supply in advance a policy for the decision at each decision variable $\Delta$. This policy usually takes the form of a specification of a decision alternative for each configuration of the parents of the decision variable. For example, in Figure 1, when the oil wildcatter must decide whether to drill or not drill, he does so knowing his earlier choice at the decision variable T as well as the value of the chance variable R.

The result of such a policy can be interpreted as a conditional for $\Delta$ given its parents. This conditional gives only probabilities of zero and one:

$$P(\Delta = d | X_1 = x_1, ..., X_k = x_k) = \begin{cases} 1 \text{ if } d = \delta(x_1,...,x_k) \\ 0 \text{ if } d \neq \delta(x_1,...,x_k) \end{cases}, \quad (2)$$

where $\delta(x_1,...,x_k)$ is the decision alternative that the decision maker deterministically chooses when he or she observes the values $x_1,...,x_k$ of the parents. For example, in Figure 1, the wildcatter may supply a policy where he would use test2, i.e., T=t2, and drill only if T=t2 and R = c.

Notice that by supplying in advance conditionals for the decision variables, the decision maker turns the ID into a BN; he or she now has conditionals for *all* the variables given their parents.

In general, a chance node X in a BN or ID is called *deterministic* if its conditional is of the type (2), where $\Delta$ is replaced by X, d is replaced by x and $\delta(x_1,...,x_k)$ is replaced by $f(x_1,...,x_k)$, f being the deterministic function that gives a value for X for every configuration $x_1,...,x_k$ of the parents. We assume that an ID has exactly one sink, which is a deterministic chance node and which has real numbers for its values. The decision maker's purpose is to choose conditionals for the decision variables so as to maximize or minimize (optimize) the expectation of this node, which is called the *value node* and designated by V.

The decision maker may, if he or she wants, adopt a policy that makes the conditional for a decision variable depend on only some or none of its parents. For example, the wildcatter may adopt a policy that makes his conditional for whether to drill or not drill based only on the value of R, but not on the value of T. In some cases, we can tell from the structure of the graph (without numerical calculations based on the conditionals for the chance nodes) that the best policy based on only some parents will do as well as the best policy based on them all. If some parents are omitted, then the corresponding arrows can be omitted; the graph with these arrows omitted will still be an ID, and once the decision maker has specified all the conditionals, it will be a BN. In general, description and computation in a BN or ID is easier with fewer arrows, so the decision maker would like to omit as many arrows as possible, but he or she may have to make his or her policy depend on all or most of the decision variables' parents in order to optimize the expectation of the value node. Thus the decision maker's objective is to make the policy depend on as few of the parents as possible while still optimizing the expectation of the value node.

Recall that the arrows into chance nodes are called relevance arrows and arrows into decision nodes are called informational arrows. Informational arrows indicate that the decision maker has certain information, not that he or she must use it. The relevance arrows will all remain in the BN that the decision maker constructs with his or her choice of conditionals for the decision variables, but the decision maker essentially omits informational arrows when he or she chooses a conditional that does not depend on the nodes from which these arrows come.

Following Clemen (1991), we distinguish between informational arrows and relevance arrows by representing the former with dashed arrows while still using solid arrows to represent the latter. Notice that decision variables are treated just like chance variables when they are parents. The difference between an informational and a relevance arrow depends on the kind of variable the arrow points to, not on the kinds of variables it comes from.

We follow Howard and Matheson (1981) and most of the ID literature in making two additional assumptions:
1. The decisions are all made by a single decision maker who remembers his or her previous decisions. This assumption is represented by the existence of a path in the DAG consisting only of all the decision variables. In other words, the decision variables are ordered, say $\Delta_1, \Delta_2,...,\Delta_k$, so that there is an arrow from $\Delta_i$ to $\Delta_{i+1}$, for i = 1, 2, ..., k-1. We summarize this by saying that the decision variables are *completely ordered* by the DAG.
2. The decision maker does not forget any previous information as he or she progresses through the decisions. This assumption is represented by the existence of an arrow from X to $\Delta_j$ whenever there is an arrow from X to $\Delta_i$ and i<j. In other words, each decision node inherits the parents of preceding decision nodes. This is called the *no-forgetting* assumption. Notice that a decision variable need not inherit the parents of chance variables that precede it. In Figure 1, for example, there is no arrow from O to D.

We conclude this section with a note on randomization that we include for completeness and that the reader may omit without loss of continuity. Recall that the decision maker can be asked to supply in advance a policy for the decision at each decision variable $\Delta$. As we discussed above this policy can take the form of a specification of a decision alternative for each configuration of the parents of the decision variable. This policy can also take the form of a probability distribution over the decision alternatives for each configuration of the parents of the decision variable. For example, in Figure 1, the wildcatter has the option of choosing a probability distribution over the decision alternatives of whether to drill or not drill. In this case a probability distribution must be specified for each configuration of D's parents. For instance, he may decide to drill with probability 1/3 given T=t1 and R=o, with probability 1/2 given T=t1 and R=c, with probability 1/5 given T=t2 and R=o, with probability 9/10 given T=t2 and R=c, and with probability 1/4 given T=nt and R=nr.

The result of such a policy can be interpreted as a conditional for $\Delta$ given its parents. Again, the decision maker has conditionals for all the variables given their parents, and the ID becomes a BN. The decision maker's objective is still to make the policy depend on as few of the parents as possible while optimizing the expectation of the value node.

However, in optimizing the expectation of the value node, there is no advantage to randomization because of the convexity of the set of probability distributions that can be attained (von Neumann and Morgenstern, 1953). Indeed, when we look at how the expectation of the value node V is affected by varying the conditional for a particular decision node $\Delta$ —fixing for the moment conditionals for the other decision variables—we see that this expectation is an average of the expectations for V that would be obtained by fixed decision functions for $\Delta$. Such an average cannot be better than the best of the expectations being averaged (Dantzig, 1951). So instead of using conditionals for decision alternatives, researchers have traditionally used decision functions (Howard and Matheson, 1981).

## 3 SOLVING INFLUENCE DIAGRAMS

Solution algorithms for IDs can all be described as elaborations of various forms of the principle of optimality in stochastic dynamic programming, which allows us to find the decision functions in problems of this type sequentially (Bellman and Dreyfus, 1962).

In its standard form, the principle of optimality in stochastic dynamic programming applies when we want to maximize or minimize the expectation of a real-valued variable V whose joint distribution with k+1 other variables $\Gamma_0, \Gamma_1, \ldots, \Gamma_k$ (which may each be vectors of variables) depends in a stagewise manner on k parameters (which also may be single numbers, vectors or functions) $\delta_1, \ldots, \delta_k$. More precisely, we assume that we can factor the joint probability for $\Gamma_0, \Gamma_1, \ldots, \Gamma_k$ and V in the form

$$P_{\delta_1,\ldots,\delta_k}(\Gamma_0,\ldots,\Gamma_k,V) = h_0(\Gamma_0)h_{\delta_1}(\Gamma_1|\Gamma_0)\ldots$$
$$\ldots h_{\delta_{k-1}}(\Gamma_{k-1}|\Gamma_0,\ldots,\Gamma_{k-2})h_{\delta_k}(\Gamma_k,V|\Gamma_0,\ldots,\Gamma_{k-1}), \quad (3)$$

where the factors are conditional probabilities. We must also assume that it is computationally feasible to compute
$$E_{\delta_k}(V|\Gamma_0,\ldots,\Gamma_{k-1})$$
from
$$h_{\delta_k}(\Gamma_k,V|\Gamma_0,\ldots,\Gamma_{k-1})$$

for each value of $\delta_k$ and each configuration of values of $\Gamma_0 \cup \ldots \cup \Gamma_{k-1}$, or at least to find for each configuration $(\gamma_0,\ldots,\gamma_{k-1})$ of $\Gamma_0 \cup \ldots \cup \Gamma_{k-1}$ the value of $\delta_k$ that optimizes

$$E_{\delta_k}(V|\Gamma_0 = \gamma_0,\ldots,\Gamma_{k-1} = \gamma_{k-1}). \quad (4)$$

Finally, we must assume (this is crucial) that we can find a single value of $\delta_k$ that optimizes (4) for all $(\gamma_0,\ldots,\gamma_{k-1})$. Since the distribution of $\Gamma_0 \cup \ldots \cup \Gamma_{k-1}$ does not depend on $\delta_k$, we have

$$E_{\delta_1,\ldots,\delta_k}(V) = E_{\delta_1,\ldots,\delta_{k-1}}\left(E_{\delta_k}(V|\Gamma_0,\ldots,\Gamma_{k-1})\right).$$

Therefore, this optimizing value of $\delta_k$ will also optimize the unconditional expectation $E_{\delta_1,\ldots,\delta_k}(V)$ for any choice of $(\delta_1,\ldots,\delta_{k-1})$. And therefore, it can be extended to a choice of $(\delta_1,\ldots,\delta_k)$ to optimize this unconditional expectation.

Suppose we fix this optimal value of $\delta_k$ eliminating it from our notation, and reducing (3) to
$$h_{\delta_1,\ldots,\delta_{k-1}}(\Gamma_0,\ldots,\Gamma_k,V) = h_0(\Gamma_0)h_{\delta_1}(\Gamma_1|\Gamma_0)\ldots$$
$$\ldots h_{\delta_{k-1}}(\Gamma_{k-1}|\Gamma_0,\ldots,\Gamma_{k-2})h(\Gamma_k,V|\Gamma_0,\ldots,\Gamma_{k-1}). \quad (5)$$

From this point, we proceed in either of two ways. We can sum or integrate $\Gamma_k$ out of the expectation. Or we can incorporate $\Gamma_k$ as part of $\Gamma_{k-1}$.

The first option, summing or integrating $\Gamma_k$ out, means reducing (5) to
$$h_{\delta_1,\ldots,\delta_{k-1}}(\Gamma_0,\ldots,\Gamma_{k-1},V) = h_0(\Gamma_0)h_{\delta_1}(\Gamma_1|\Gamma_0)\ldots$$
$$\ldots h'_{\delta_{k-1}}(\Gamma_{k-1},V|\Gamma_0,\ldots,\Gamma_{k-2}),$$
where

$$h'_{\delta_{k-1}}(\Gamma_{k-1},V|\Gamma_0,\ldots,\Gamma_{k-2}) =$$
$$h_{\delta_{k-1}}(\Gamma_{k-1}|\Gamma_0,\ldots,\Gamma_{k-2})\int h(\gamma_k,V|\Gamma_0,\ldots,\Gamma_{k-1})d\gamma_k.$$

Once again, we assume that we can choose a single value of $\delta_{k-1}$ so as to optimize simultaneously

$$E_{\delta_{k-1}}(V|\Gamma_0 = \gamma_0,\Gamma_1 = \gamma_1,\ldots,\Gamma_{k-2} = \gamma_{k-2}) \quad (6)$$

for all $(\gamma_0,\ldots,\gamma_{k-2})$. Then, as before, the choice of $\delta_{k-1}$ can be extended to a choice of $(\delta_1,\ldots,\delta_{k-1})$ to optimize the unconditional expectation

$$E_{\delta_1,\ldots,\delta_{k-1}}(V).$$

So we may also fix this optimal value of $\delta_{k-1}$, and reduce the problem further. We can continue in this way, choosing the $\delta_i$ sequentially, provided that the successive simultaneous optimizations like those in (4), (6), etc. are possible.

The second option means setting $\Gamma'_{k-1} = \Gamma_{k-1} \cup \Gamma_k$, and reducing (5) to

$$h_{\delta_1,\ldots,\delta_{k-1}}(\Gamma_0,\ldots,\Gamma_{k-1},V) = h_0(\Gamma_0)h_{\delta_1}(\Gamma_1|\Gamma_0)\ldots$$
$$\ldots h'_{\delta_{k-1}}(\Gamma'_{k-1},V|\Gamma_0,\ldots,\Gamma_{k-2}),$$
where

$$h'_{\delta_{k-1}}(\Gamma'_{k-1},V|\Gamma_0,\ldots,\Gamma_{k-2}) =$$
$$h_{\delta_{k-1}}(\Gamma_{k-1}|\Gamma_0,\ldots,\Gamma_{k-2})h(\Gamma_k,V|\Gamma_0,\ldots,\Gamma_{k-1}).$$

Again, if we can choose $\delta_{k-1}$ to optimize simultaneously (6) for all $(\gamma_0,\ldots,\gamma_{k-2})$, and so on, we can proceed to choose the $\delta_i$ sequentially.

This standard version of stochastic dynamic programming is not quite adequate for the case of influence diagrams. The reason is that although these diagrams involve factorizations that can be written in the form (3), the factors are not necessarily conditional probabilities.

The standard version of stochastic dynamic programming can be modified to fit influence diagrams, but there has been a considerable variety of opinion about how to do this. The oldest sequential solution algorithm for influence diagrams, the Olmsted-Shachter reduction algorithm (Olmsted, 1983; Shachter, 1986) goes considerably beyond stochastic dynamic programming, in order to maintain a representation of the influence diagram form as the algorithm proceeds. More recent algorithms, including the valuation network algorithm of Shenoy (1992 and 1993) and the potential influence diagram algorithm of Ndilikilikesha (1992), stay closer to stochastic dynamic programming.

The simulation algorithm we will describe in Section 5, does not fit exactly into either Shenoy's or Ndilikilikesha's framework, primarily because their algorithms integrate $\Gamma_k$ out, while our algorithm follows the second option described above, that of absorbing $\Gamma_k$ into $h_{\delta_{k-1}}$. We could elaborate one of their frameworks in order to make our algorithm fit, but it will be simpler for us to deal directly with the necessary modification in the standard form of stochastic dynamic programming that we have just described.

Here is the modification that we require. Let us assume that the joint probability for $\Gamma_0, \Gamma_1, ..., \Gamma_k$ and $V$ is proportional to a factorization of the following form

$$P_{\delta_1,...,\delta_k}(\Gamma_0,...\Gamma_k, V) \propto h_0(\Gamma_0) h_{\delta_1(\Gamma_0)}(\Gamma_1|\Gamma_0)...$$
$$...h_{\delta_{k-1}(\Gamma_0,...,\Gamma_{k-2})}(\Gamma_{k-1}|\Gamma_0,...,\Gamma_{k-2})\cdot$$
$$\cdot h_{\delta_k(\Gamma_0,...,\Gamma_{k-1})}(\Gamma_k, V|\Gamma_0,...,\Gamma_{k-1}) \quad (7)$$

Here we do **not** assume that the factors are conditional probabilities. But we **do** assume that the $\delta_i$ are functions; and we assume, as the notation indicates, that for fixed values of $\Gamma_0, \Gamma_1, ..., \Gamma_{k-1}$, the factor $h_{\delta_k(\Gamma_0,...,\Gamma_{k-1})}$, regarded as a function of $\Gamma_k$ and $V$, depends on $\delta_k$ only through the value $\delta_k$ assigns to those values of $\Gamma_0, \Gamma_1, ..., \Gamma_{k-1}$. This assumption, as we will see implies that the simultaneous optimizations at each step are possible.

Notice first that the factorization (7) implies that

$$h_{\delta_k(\Gamma_0,...,\Gamma_{k-1})}(\Gamma_k, V|\Gamma_0,...,\Gamma_{k-1}),$$

for fixed values of $\Gamma_0, \Gamma_1, ..., \Gamma_{k-1}$, is at least proportional to the conditional probability distribution for $\Gamma_k$ and $V$ given these values of $\Gamma_0, \Gamma_1, ..., \Gamma_{k-1}$. To see this, recall that a conditional probability distribution is always proportional to the corresponding joint probability distribution. Thus

$$P_{\delta_1,...,\delta_k}(\Gamma_k, V|\Gamma_0,...,\Gamma_{k-1}) = \lambda P_{\delta_1,...,\delta_k}(\Gamma_0,...\Gamma_k, V), (8)$$

where $\lambda$ is constant with respect to $\Gamma_k$ and $V$. (The other variables are thought of as fixed.) We usually write (8) with a symbol of proportionality:

$$P_{\delta_1,...,\delta_k}(\Gamma_k, V|\Gamma_0,...,\Gamma_{k-1}) \propto P_{\delta_1,...,\delta_k}(\Gamma_0,...\Gamma_k, V).$$

Since only the last factor of (7) involves $\Gamma_k$ or $V$, (8) implies that

$$P_{\delta_1,...,\delta_k}(\Gamma_k, V|\Gamma_0,...,\Gamma_{k-1}) \propto$$
$$h_{\delta_k(\Gamma_0,...,\Gamma_{k-1})}(\Gamma_k, V|\Gamma_0,...,\Gamma_{k-1}).$$

Again, this proportionality is to be interpreted by taking both sides as functions of $\Gamma_k$ and $V$ only, with the other variables fixed; we are able to omit the other factors only because they, as functions of the other variables, are also fixed and hence can be absorbed into the constant of proportionality.

Whenever a function is proportional to a probability distribution (or probability conditional), it contains all the information needed to find that conditional because the constant of proportionality is simply what is needed to make the function sum (or integrate) to one. Thus

$$h_{\delta_k(\gamma_0,...,\gamma_{k-1})}(\Gamma_k, V|\Gamma_0 = \gamma_0,...,\Gamma_{k-1} = \gamma_{k-1})$$

has, in particular, all the information needed to determine the conditional expectation of $V$ given $(\gamma_0,...,\gamma_{k-1})$,

$$E_{\delta_k(\gamma_0,...,\gamma_{k-1})}(\Gamma_k, V|\Gamma_0 = \gamma_0,...,\Gamma_{k-1} = \gamma_{k-1}). \quad (9)$$

We can choose the value of $\delta_k(\gamma_0,...,\gamma_{k-1})$ to optimize this expectation, and by doing this for each set of values $(\gamma_0,...,\gamma_{k-1})$, we will have chosen a function $\delta_k$ that simultaneously optimizes (9) for all $(\gamma_0,...,\gamma_{k-1})$.

Once this choice of $\delta_k$ has been carried out, we can proceed, as before, absorbing $h_{\delta_k(\Gamma_0,...,\Gamma_{k-1})}$ into $h_{\delta_{k-1}(\Gamma_0,...,\Gamma_{k-2})}$, first integrating $\gamma_k$ out if we wish to do so. This means replacing the factor in the second line of (7) with a factor

$$h'_{\delta_{k-1}(\Gamma_0,...,\Gamma_{k-2})}(\Gamma'_{k-1}, V|\Gamma_0,...,\Gamma_{k-2}),$$

where

(1) $\Gamma'_{k-1} = \Gamma_{k-1} \cup \Gamma_k$,
(2) $\Delta_k$, which is in $\Gamma_k$, is now interpreted as a chance node, and
(3) $h'_{\delta_{k-1}(\Gamma_0,...,\Gamma_{k-2})}(\Gamma'_k, V|\Gamma_0,...,\Gamma_{k-2}) =$
$$h_{\delta_{k-1}(\Gamma_0,...,\Gamma_{k-2})}(\Gamma_{k-1}|\Gamma_0,...,\Gamma_{k-2})\cdot$$
$$\cdot h^*_{\Delta_k}(\Delta_k|\Gamma_0,...,\Gamma_{k-1})h_{\delta^*_k}(\Gamma_k, V|\Gamma_0,...,\Gamma_{k-1}),$$

where $\delta^*_k$ is the optimal decision function and $h^*_{\Delta_k}$ is the zero-one conditional representing $\delta^*_k$.

In order to fit influence diagrams into this version of stochastic dynamic programming, we write $\Gamma_i$ for the set of variables consisting of $\Delta_i$ together with the chance variables observed by the decision maker between $\Delta_i$ and $\Delta_{i+1}$, for $i = 1, ..., k-1$, we write $\Gamma_0$ for the chance variables observed before $\Delta_1$ and $\Gamma_k$ for the set of variables consisting of $\Delta_k$ together with the chance variables (other than V) observed after $\Delta_k$ (or never), and we write $\delta_i$ for the decision function for $\Delta_i$. Then we set $h_0(\Gamma_0)$ equal to the product of conditionals for the chance variables in $\Gamma_0$. For $i = 1, ..., k-1$, we set $h_{\delta_i}(\Gamma_i|\Gamma_0,...,\Gamma_{i-1})$ equal to the product of conditionals for the chance variables in $\Gamma_i$, times the conditional corresponding to the decision function $\delta_i$ (recall Formula (2)). And we similarly set $h_{\delta_k}(\Gamma_k,V|\Gamma_0,...,\Gamma_{k-1})$ equal to the product of the conditionals for all variables in $\Gamma_k \cup \{V\}$. Since $h_{\delta_k}(\Gamma_k,V|\Gamma_0,...,\Gamma_{k-1})$ depends on $\delta_k$ only through its value $\delta_k(\Gamma_0,...,\Gamma_{k-1})$, this puts us in the framework just described.

In the computational theory of Section 5 we will use Gibbs sampling to implement the method just described.

## 4 INFORMATION/RELEVANCE INFLUENCE DIAGRAMS

In the standard influence diagram, arrows into a decision node are only informational; they do not represent constraints on what the decision maker can do. We can represent such constraints only indirectly, using arrows to the children of the decision.

Users of influence diagrams often want to represent constraints by arrows into decision nodes. For example, a user might draw Figure 2 in an attempt to represent the fact that the budget, B, constrains the options for testing and drilling. However, Figure 2 cannot represent such constraints because the dashed arrows are only informational, and the figure does not show B having any relevance to any chance variables. If Figure 2 is interpreted as a standard influence diagram, then variable B will have absolutely no effect on the optimal decision functions; they are the same with or without it.

**Figure 2.** ID for the oil wildcatter's problem with budget constraint as information only

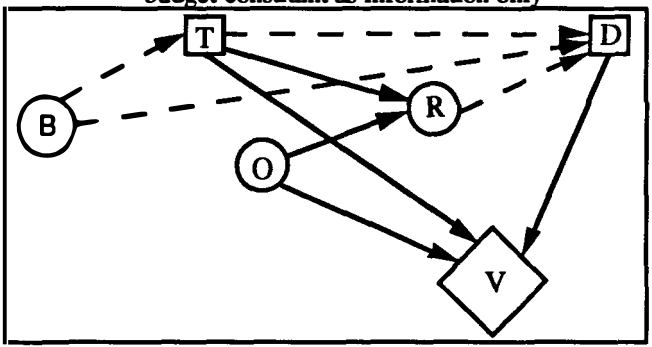

In order to represent a budget constraint using the standard ID representation, we would have to complicate the diagram of Figure 2 in some way. One way is represented in Figure 3. Here D' is a deterministic chance node representing whether drilling actually takes place as a function of the budget and the decision maker's decisions. For example, if the possible budget amounts are $1M (million) or $2M (million), the tests costs are $50,000 for test1 and $100,000 for test2, and the cost of drilling is $950,000, then the decision maker cannot drill if B = $1M and T = t2. Using Figure 3, we can represent this by making D' (whether the drilling really takes place) a deterministic function of B, T and D as follows:

(1) D' = f(B = $2M, T = nt, D = d) = yes
(2) D' = f(B = $2M, T = nt, D = nd) = no
(3) D' = f(B = $2M, T = t1, D = d) = yes
(4) D' = f(B = $2M, T = t1, D = nd) = no
(5) D' = f(B = $2M, T = t2, D = d) = yes
(6) D' = f(B = $2M, T = t2, D = nd) = no
(7) D' = f(B = $1M, T = nt, D = d) = yes
(8) D' = f(B = $1M, T = nt, D = nd) = no
(9) D' = f(B = $1M, T = t1, D = d) = yes
(10) D' = f(B = $1M, T = t1, D = nd) = no
(11) D' = f(B = $1M, T = t2, D = d) = no
(12) D' = f(B = $1M, T = t2, D = nd) = no

As line (11) indicates, the decision maker's purported decision to drill does not result in drilling if he or she does not have the money.

**Figure 3.** ID for the oil wildcatter's problem with budget constraint having an effect on the decision function for D

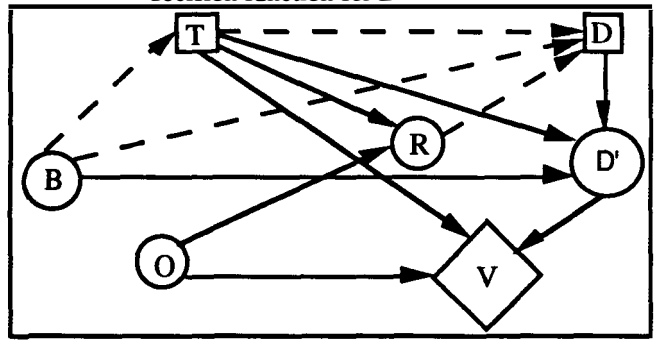

The complexity of Figure 3 is obviously undesirable. This suggests that we go beyond the standard ID definitions and allow relevance arrows into decision nodes. Such arrows would indicate both that the decision maker knows the value of the variables from which these arrows emanate when he or she makes the decision, and *also* that the variables constrain the decision.

This suggests the following formal definitions.

**Definition 2.** A *constraint* on a variable X given a set of variables $X_1, ..., X_k$ is a mapping $C_X$ from the frame of $X_1, ..., X_k$ to subsets of the frame of X. (In other words, for each configuration $x_1, ..., x_k$ of $X_1, ..., X_k$, we specify a set $C_X(x_1, ..., x_k)$ of permitted values for X.)

**Definition 3.** An *information/relevance influence diagram (IRID)* is a DAG with variables as nodes, some of which are called *chance* variables and some of which are called *decision* variables, together with
(1) a specification, for each chance node X, of a conditional for X given its parents, and
(2) a specification, for each decision node $\Delta$, of a constraint for $\Delta$ given a subset $S_\Delta$ of its parents.

We divide the arrows into *relevance* arrows, which are solid, and *informational* arrows, which are dashed, as follows:
(1) all arrows into a chance node X are relevance arrows,
(2) arrows into a decision node Δ from parents in $S_\Delta$ are relevance arrows, and
(3) all other arrows into Δ are informational arrows.

Here, as in the case of IDs, we assume that there is a *value* node: a deterministic chance node which is a sink and is real-valued.

We interpret IRIDs by assuming that when the decision maker makes decision Δ, he or she has observed *all* the parents of Δ, including both those from which there are informational arrows and those from which there are relevance arrows. The relevance arrows indicate information for the decision maker as well as constraints on the decision.

We assume complete ordering and no-forgetting for decision variables:
(1) There is a path in the DAG consisting only of decision variables. (Some or all of the arrows on this path may be relevance arrows.)
(2) If the decision variables, $\Delta_1, ..., \Delta_k$, are ordered by the path that joins them and there is an arrow from X to $\Delta_i$, then there is an arrow from X to $\Delta_j$, for all i<j. (Again, there is no restriction on whether these arrows are relevance or informational arrows, or whether one is a relevance arrow and the other is an informational arrow.)

Figure 4 shows how the budget constraint for the wildcatter problem can be represented in an IRID. We start with the ID of Figure 2. We create a chance node for the variable B. Then we draw an informational arrow from B to T, a relevance arrow from B to D, and we replace the informational arrow from T to D by a relevance arrow.

**Figure 4.** IRID for the oil wildcatter's problem

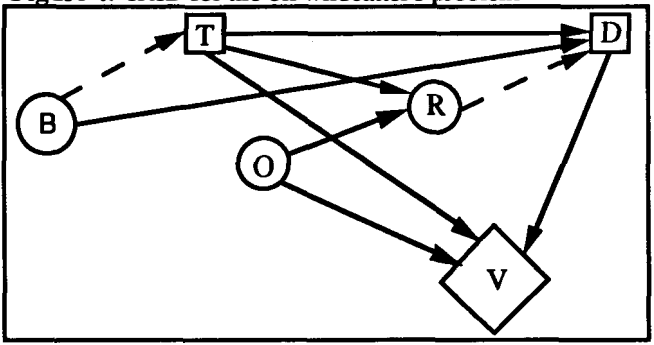

For the diagram of Figure 4 we specify the following conditionals and constraints:
P(B = $1M)=0.50=1 − P(B = $2M)
P(O = w)=0.60=1 − P(O = y)
P(R=o | O=w, T=t1)=0.20=1−P(R=c | O=w, T=t1)
P(R=o | O=w, T=t2)=0.05=1−P(R=c | O=w, T = t2)
P(R=o | O=y, T=t1)=0.90=1−P(R=c | O=y, T=t1)
P(R=o | O=y, T=t2)=0.90=1−P(R=c | O=y, T=t2)
P(R = nr | O, T = nt) = 1.0
V(O = w, T = nt, D = nd) = −$2,000,000
V(O = y, T = nt, D = nd) = $0
V(O = w, T = nt, D = d) = $1,050,000
V(O = y, T = nt, D = d) = −$950,000
V(O = w, T = t1, D = nd) = −$2,050,000
V(O = y, T = t1, D = nd) = −$50,000
V(O = w, T = t1, D = d) = $1,000,000
V(O = y, T = t1, D = d) = −$1,000,000
V(O = w, T = t2, D = nd) = −$2,100,000
V(O = y, T = t2, D = nd) = −$100,000
V(O = w, T = t2, D = d) = $950,000
V(O = y, T = t2, D = d) = −$1,050,000
$C_D$(B = $1M, T = nt) = {d, nd}
$C_D$(B = $1M, T = t1) = {d, nd}
$C_D$(B = $1M, T = t2) = {nd}
$C_D$(B = $2M, T = nt) = {d, nd}
$C_D$(B = $2M, T = t1) = {d, nd}
$C_D$(B = $2M, T = t2) = {d, nd}

Notice that the theory of stochastic dynamic programming that applies to IDs also applies to IRIDs; it is simply necessary that each step in the optimization respect the constraints.

## 5 SOLVING IRIDS

We begin this section with a review of the simulation algorithm for solving IDs given in Jenzarli (1995). Then we adopt the algorithm to IRIDs. Finally, we apply the adapted algorithm to the IRID of the oil wildcatter problem.

Jenzarli (1995) shows how to use Gibbs sampling (Geman and Geman, 1984; and Gelfand and Smith, 1990), an iterative Markov chain Monte Carlo algorithm (Hastings, 1970), to implement stochastic dynamic programming for an influence diagram. For an ID with k decision variables $\Delta_1,...,\Delta_k$ and sets $\Gamma_0,...,\Gamma_k$ as described in Section 3, we give a brief summary of the simulation algorithm.

Our task is to find the decision function $\delta_k$. This means finding, for each configuration $(\gamma_0, \gamma_1,...,\gamma_{k-1})$ of $\Gamma_0 \cup ... \cup \Gamma_{k-1}$, the value $d_k$ of the decision $\Delta_k$ that optimizes

$$E_{d_k}(V|\Gamma_0 = \gamma_0, \Gamma_1 = \gamma_1,...,\Gamma_{k-1} = \gamma_{k-1}). \quad (10)$$

(Notice that we write $d_k$ in the place of $\delta_k$ as a subscript on the expectation operator; this is because the expectation for the configuration $(\gamma_0, \gamma_1,...,\gamma_{k-1})$ of the predecessors depends only on the value $d_k$ that $\delta_k$ assigns to this configuration.) To this end, we simply compute (10) for all $d_k$ and choose the $d_k$ that gives the optimal (largest or smallest depending on whether we are maximizing or minimizing) result.

336 Jenzarli

To compute (10) for a particular $d_k$, we recall that the conditional joint distribution of $\Gamma_k \cup \{V\}$ is proportional to $h_{d_k}(\Gamma_k, V | \Gamma_0 = \gamma_0, ..., \Gamma_{k-1} = \gamma_{k-1})$, which is simply the product of the conditionals for $\Gamma_k \cup \{V\}$. We now give the steps involved in the computation of (10).

(1) Delete all barren nodes form the ID (Shachter, 1986).
(2) Draw the ID according to the information partitioning implied by $\Gamma_0, ..., \Gamma_k$ and omit informational arrows.
(3) Form the moral graph.
(4) Omit any variables from $\Gamma_k$ that are not connected with V in the subgraph of this moral graph determined by $\Gamma_k \cup \{V\}$. The result is $\Gamma'_k$.
(5) Omit any variables not in $\Gamma'_k \cup \{V\}$ that are not neighbors of $\Gamma'_k$ in the moral graph of Step 4. The variables that remain are the variables on which $\Delta_k$ depends that are not in $\Gamma'_k \cup \{V\}$.
(6) Draw the directed subgraph determined by all the variables that remain.
(7) The factors remaining in the joint distribution are those in which the variables remaining in $\Gamma'_k$ are either parents or children.
(8) Now that the factors are identified, we do Gibbs sampling with these factors to simulate the joint distribution of $\Gamma'_k \setminus \{\Delta_k, V\}$. First, we fix the variables on which $\Delta_k$ depends that are not in $\Gamma'_k \cup \{V\}$. Then, for the configuration of $\Gamma'_k \setminus \{\Delta_k, V\}$ obtained at each step of the Gibbs sampling, we compute V. This gives a sequence of values for V simulating a random sample from its conditional distribution, from which we may compute its conditional expectation.
(9) When we move on to the next step of the stochastic dynamic program, we use the second of the two options discussed in Section 3. In other words, we absorb $\Gamma_k$ into $\Gamma_{k-1}$, and we include the conditionals from $\Gamma_k$ in the new factorization of $h_{\delta_{k-1}}$. In order to avoid zero probabilities that would interfere with the Gibbs sampling, we do not include the conditional for $\Delta_k$ corresponding to the decision function we have just found for $\Delta_k$. Instead, we substitute this decision function in all the conditionals in which $\Delta_k$ appeared as a parent, thus eliminating $\Delta_k$ from the graph and producing arrows from the variables on which $\Delta_k$ depends to the variables for which it was a parent.

The algorithm we have just described for IDs also works for IRIDs. The only point to note is that the relevance arrows into $\Delta_k$ must be included in the graph from which the moral graph is formed. We now apply the algorithm to the IRID of the oil wildcatter problem. Figure 5 shows the IRID of the oil wildcatter problem with informational arrows omitted, where
$\Gamma_0 = \{B\}$, $h_0(B) = P(B)$,
$\Delta_1 = T$, $\Gamma_1 = \{T, R\}$, $h_{\delta_1}(T, R | B) = P(T) P(R | T, O)$,
$\Delta_2 = D$, $\Gamma_2 = \{D, O\}$, and
$h_{\delta_2}(D, O, V | B, T, R) = P(D | T, B) P(O) P(V | T, D, O)$.

**Figure 5.** IRID of the oil wildcatter problem with informational arrows omitted

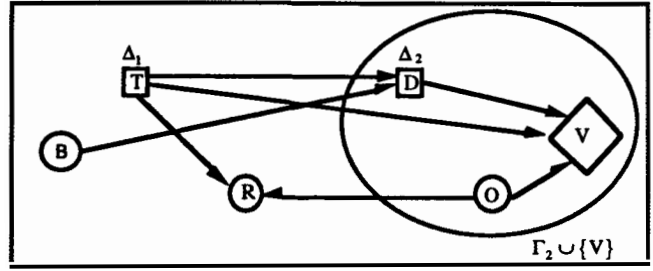

We now form the moral graph as shown in Figure 6. Notice that according to Step 4 of the algorithm, $\Gamma_2 = \Gamma'_2$. Also, all variables not in $\Gamma'_2 \cup \{V\}$ are neighbors of $\Gamma'_2$ in the moral graph of Figure 6. This means that the variables on which D depends are T, R, and B. Thus the factors to consider in the joint distribution are those in which D and O are either parents or children. These factors on which we do Gibbs sampling are:

$P(R = \bullet | T = \bullet, O) P(D = \bullet | T = \bullet, B = \bullet) P(O)$,

where $\bullet$ means that the value of the variable in question is fixed.

**Figure 6.** Moral graph for the IRID of Figure 5

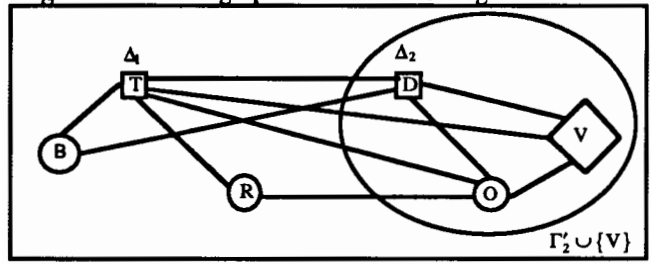

Next we do Gibbs sampling with these factors to simulate the posterior distribution of O. First, we fix the variables on which D depends, i.e., T, R, and B. Then for every value of O, we compute V. This gives a sequence of values for V simulating a random sample from its conditional distribution, from which we may compute its conditional expectation. We perform this step for every value of D, and we choose the value of D that corresponds to the maximum conditional expectation of V.

When we move on to the next step of the stochastic dynamic program, we absorb $\Gamma_2$ into $\Gamma_1$ as described in Step 9 of the algorithm. Then we eliminate D from the graph, thus producing arrows from T, R, and B to V. The resulting graph is shown in Figure 7.

**Figure 7.** DAG with D absorbed into its direct successors

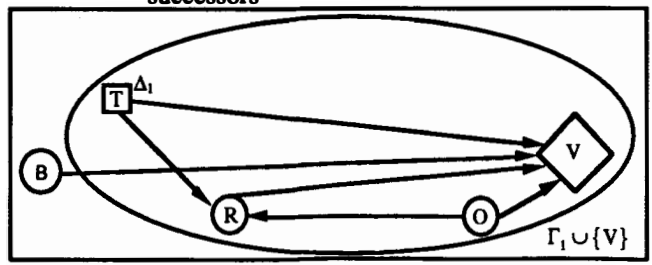

Finally, we include the conditionals from $\Gamma_2$ in the new factorization of $h_{\delta_1}$. In order to avoid zero probabilities that would interfere with the Gibbs sampling, we do not include the conditional for D corresponding to the decision function we have just found for D. Instead, we substitute this decision function in all the conditionals in which D appeared as a parent.

### Acknowledgments

This work was partially supported by the National Science Foundation under grants IRI–8902444 and SES–9213558, and by a grant from the General Research Fund at the University of Kansas, grant 3605–XX–0038. Other partial support came from a Globe Project grant at the University of Tampa. I am grateful for comments from my mentors Glenn Shafer and Prakash Shenoy. I am especially grateful for the help and encouragement I received from Glenn Shafer.

### References

Bellman, R. and Dreyfus, S. (1962) *Applied Dynamic Programming*, Princeton University Press, Princeton, N.J.

Clemen, R.T. (1991) *Making Hard Decisions*, Duxbury Press, Belmont, California.

Gelfand, A. E. and Smith, A. F. M. (1990) Sampling-based approaches to calculating marginal densities, *Journal of the American Statistical Association*, 85(410), 398-409.

Geman, S. and Geman, D. (1984) Stochastic relaxation, Gibbs distributions, and the Bayesian restoration of images, *IEEE Transactions on Pattern Analysis and Machine Intelligence*, PAMI-6, 721-741.

Hastings, W. K. (1970) Monte Carlo sampling methods using Markov chains and their applications, *Biometrika*, 57(1), 97-109.

Howard, R. A. and Matheson, J. E. (1981) Influence diagrams. In R. A. Howard and J. E. Matheson (eds.), *Readings on the principles and Applications of Decision Analysis*, vol. 2, Strategic Decisions group, Menlo Park, CA, 1984, 719-762.

Jensen, F. V., Lauritzen, S. L., and Olesen, K. G. (1990) Bayesian updating in causal probabilistic networks by local computations, *Computational Statistics Quarterly* 4, 269-282.

Jenzarli, A. (1995) Solving Influence Diagrams Using Gibbs Sampling, *Proceedings of the Fifth International Workshop on Artificial Intelligence and Statistics*, 278-284.

Ndilikilikesha, P. (1992) Potential influence diagrams, *Working Paper No. 235*, School of Business, University of Kansas, Lawrence, Kansas.

Olmsted, S. M. (1983) On representing and solving decision problems, Ph.D. thesis, Department of Engineering-Economic Systems, Stanford University, Stanford, CA.

Pearl, J. (1988) *Probabilistic Reasoning in Intelligent Systems: Networks of Plausible Inference*, Morgan Kaufman, San Mateo, CA.

Raiffa, H. (1968) *Decision Analysis*, Addison-Wesley, Reading, MA.

Shachter, R. D. (1986) Evaluating influence diagrams, *Operations Research*, 34(6), 871-882.

Shenoy, P. P. (1992) Valuation-based systems for Bayesian decision analysis, *Operations Research*, 40(3), 463-484.

Shenoy, P. P. (1993) A new method for representing and solving Bayesian decision problems, in D. J. Hand (ed.), *Artificial Intelligence Frontiers in Statistics: AI and Statistics III*, 119-138, Chapman & Hall, London.

Shenoy, P. P. (1993) Valuation Network Representation and Solution of Asymmetric Decision Problems, *Working Paper No. 246*, School of Business, University of Kansas, Lawrence, Kansas.

Smith, J. E., Holtzman, S. and Matheson, J. E. (1993) Structuring Conditional Relationships in Influence Diagrams, *Operations Research*, 41(2), 280-297.